\documentclass[conference,a4paper]{IEEEtran}
\IEEEoverridecommandlockouts
\usepackage{amsmath,amssymb,amsfonts}
\usepackage{algorithmic}
\usepackage{graphicx}
\usepackage{textcomp}
\usepackage{xcolor}
\usepackage{cite}
\usepackage{subfigure}
\begin{document}
\title{Label or Message: A Large-Scale Experimental Survey of Texts and Objects Co-Occurrence }

\author{\IEEEauthorblockN{Koki Takeshita,
Juntaro Shioyama, Seiichi Uchida}
\IEEEauthorblockA{Kyushu University, Fukuoka, Japan}
}

\maketitle

\begin{abstract} 
Our daily life is surrounded by textual information. Nowadays, the automatic collection of textual information becomes 
possible owing to the drastic improvement of scene text detectors and recognizer. The purpose of this paper is to conduct a large-scale survey of co-occurrence between  visual objects (such as book and car) and scene texts 
with a large image dataset and a state-of-the-art scene text detector and recognizer. Especially, we focus on the function of ``label'' 
texts, which are attached to objects for detailing the objects. By analyzing co-occurrence between objects and scene texts, it is possible to observe the statistics about the label texts and understand how the scene texts will be useful 
for recognizing the objects and vice versa.
\end{abstract}


% \fi
\IEEEpeerreviewmaketitle

%================================================================
\section{Introduction}
\begin{figure}[t]
\centering
\includegraphics[width=0.45\textwidth]{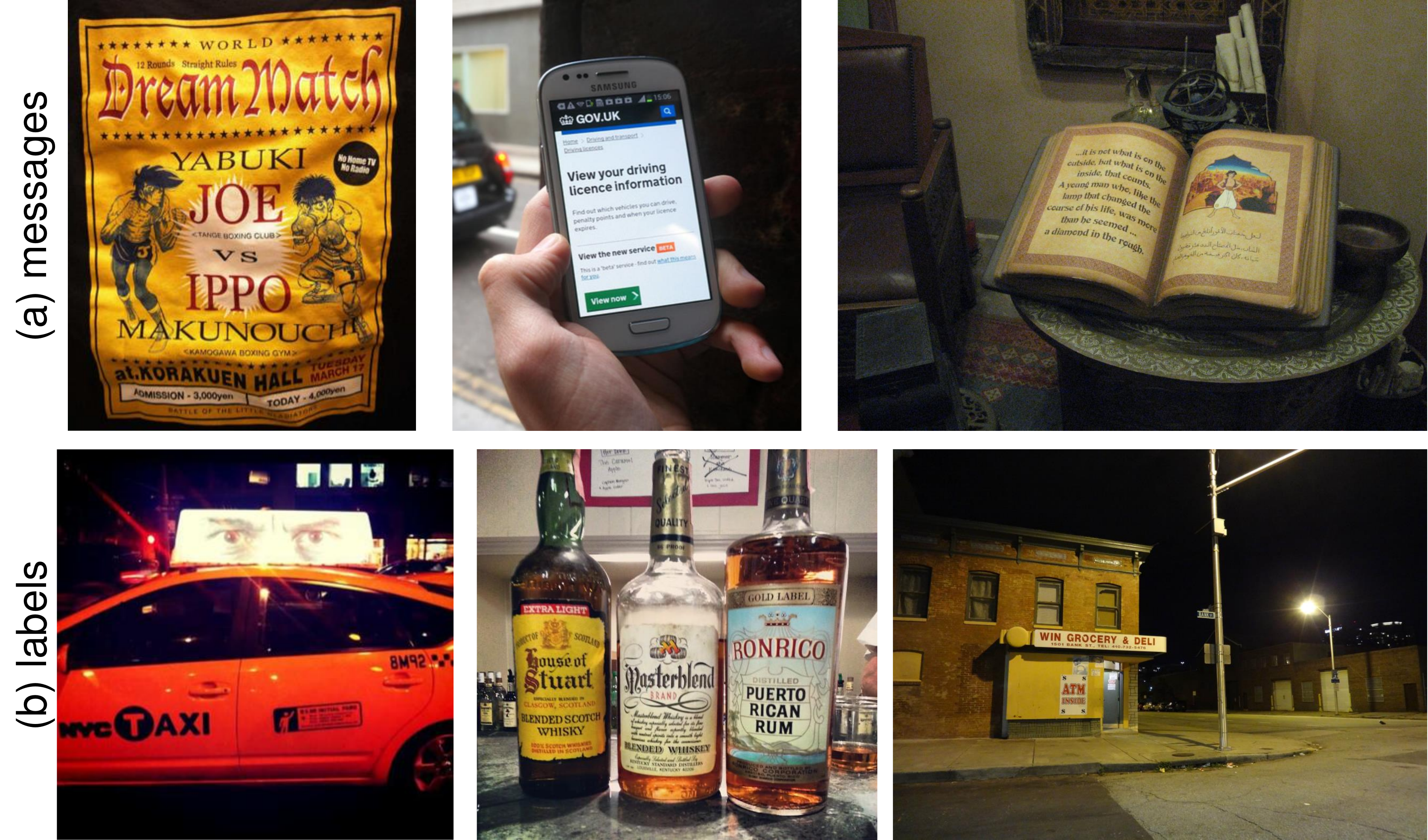}\\[-3mm]
\caption{Messages and labels.}
\label{fig:label_message_word}
\end{figure}

%================================================================
The roles of textual information around us can be roughly divided into two types: message and label. Message texts are used for linguistic communication between humans. Texts on book pages, display screens, and notebooks are messages. Book pages (i.e., papers) and display screens are just media for carrying messages and independent of the message content. In other words, the value of a message shown in a book page will not change even if the message is shown on a smartphone display. Fig.~\ref{fig:label_message_word}~(a) shows examples of messages. The content of a poster is a message and the poster paper is a medium, i.e., a message-carrier. The texts on the smartphone display do not describe the smartphone itself and therefore they are messages.\par
Label texts are used for linguistic communication between humans and objects. The role of label texts, therefore, is the disambiguation of the object to which the text is attached. For example, if a bottle has no textual information on it, we cannot understand what is inside it. Similarly, if a building has no signboard, we also cannot understand what business is carried on there. Fig.~\ref{fig:label_message_word}~(b) shows examples of label texts. We can readily understand that this car is a taxi by the word ``taxi'' on the car body. The paper labels on the bottles, as the name implies, act as the labels for clarifying the contents of the bottles.\par
If we can detect and recognize label texts on objects, it is possible to utilize the label information for recognizing the objects with higher accuracy and/or finer-grained classes. Assume an object printed with a ``school bus.''  We can recognize it correctly as a bus even though its appearance resembles a train. Moreover, we can recognize it not just a bus but as a school bus. Like this case, label texts on an object will be very useful for object recognition. Note that it is also true that  
the object class will be useful to recognize label texts --- if a text ``bvs'' is recognized on a bus, we can correct it as ``bus.''\par
The main purpose of this paper is to provide a large-scale experimental survey on the interactions between objects and texts (words) on the objects as fundamental statistical data for future research that utilizes scene texts. 
For this purpose, a state-of-the-art scene text detector and a recognizer are first applied to the about 1.7 million images provided as OpenImages v4. Then, the co-occurrence between the recognized words 
and specific objects is evaluated. Here, co-occurrence between a specific word (e.g., ``taxi'') and an object (e.g., car) is defined as the situation that the word bounding box is completely surrounded by the object bounding box.\par
Using the co-occurrence, various analyses can be made to have a deeper understanding of object-word interaction. For example, we measure the word variations on a specific object as an entropy value --- if an object has a larger entropy, this means the words on it have a large variation and therefore the object tends to be a message-carrier. In addition, if a word has a smaller entropy, the word occurs only with a specific object and therefore the word is useful for object recognition.
\par
To the authors' best knowledge, there is no fundamental survey of the object-word co-occurrence so far, although recent research has started to utilize the co-occurrence as the interactions between objects and scene texts, as we will see in the next section. This simple but important experimental survey, therefore, will be useful to understand the reason why recent research could achieve successful results. Moreover, it will lead to new research directions that utilize object-word interaction.\par
The main contributions of our experimental survey are multifold.  
\begin{itemize}
    \item We provide the list of the frequent words captured in real images and the list of the objects where words are frequently printed.
    \item We observe object-word pairs with high co-occurrences.  
    \item We evaluate the object variations of a specific word as an entropy value and then understand what kind of words can be a label, which is useful information to improve object recognition accuracy. 
    \item Similarly, we evaluate the word variations on a specific object class as an entropy value for understanding what kind of objects can be a message-carrier. 
\end{itemize}

%================================================================
\section{Related Work}
%================================================================
%for example, Places365~\cite{zhou} achieves more than 50\% accuracy using CNN, however, it doesn't attempt to extract and use character information separately.
%
\subsection{Fine-Grained Object Recognition Using Texts\label{sec:fine-grained}}
Although the label texts on objects seem to be very useful to recognize the objects, conventional object recognition methods provide the object class without using text information. One of its reasons is that text detection and recognition in scenery (and other real) images were still difficult until the mid-2010s. Nowadays, scene text detection and recognition performance is greatly improved by convolutional neural networks (CNN), 
and we can find several attempts to perform fine-grained image classification using scene text information~\cite{context0, volkmar, yair, context1, context2, bai}.  For example, Movshovitz-Attias et al.~\cite{yair} show that scene text information (especially information printed on signboards) is effective for storefronts classification. In~\cite{context2}, recognition experiments on 28 building classes (such as Bookstore and Theatre) of ImageNet are also conducted. Even though the text recognition accuracy was not high (28.4 mAP), the paper published in 2017 proves that the text information still can improve the image-based classification accuracy (60.3$\to$70.7 mAP). 
Bai et al.~\cite{bai} also integrate scene text information and the image information for classifying the same 28-class building image dataset and a 20-class drink bottle dataset.
A recent trial by Dey et al.~\cite{dey} tries to understand the advertisement images by utilizing the textual information along with the visual contents in the images. 
\par
\subsection{Text Detection and Recognition Using Objects}
Although the above attempts utilize scene text information for recognizing and understanding the surrounding visual objects, it is also possible to utilize the object information for detecting and recognizing scene texts. In fact, the object type will give a good prior for scene text detection and recognition~\cite{kunishige, anna}. For example, several texts may be attached to the object ``buildings,'' whereas they are not to the object ``tree.'' Our analysis results (shown as Tables~\ref{table:object_freq_words} and \ref{table:object_less_freq_words}) will be useful to improve those text detection techniques utilizing the object information.
\subsection{Scene Text VQA}
In 2019, a new visual question answering (VQA) task, called 
scene text VQA, is proposed and tackled by several groups~\cite{vqa1,vqa2,vqa3,vqa4}. The questions of the task cannot be answered without reading the textual information in the scenery image. For example, the task is to answer questions such as ``Where is the destination of the blue bus'' and ``How much is the price of carbonated water?'' while watching the whole image. In the former case, after finding a blue bus by image recognition, it is necessary to detect and recognize the destination information written in front of the bus. In the latter case, it is necessary to first find the bottle by image recognition and then find and read the price label nearby the bottle. Recent technologies already start handling such complicated problems. The dataset for this task has also been released and a competition was held at ICDAR2019~\cite{vqa-competition}. 
\par

\subsection{Techniques Required for Our Experimental Survey}
The reliability of our experimental survey relies on the 
performance of the scene text detection and recognition task, because OpenImages has no ground-truth of text bounding boxes. In the past, the task was very difficult; nowadays, however, as noted in Section~\ref{sec:fine-grained} it already arrives at a practical level by the efforts on creating annotated large-scale datasets and developing deep learning technologies. Enormous methods have been proposed even in recent years and Baek et al.~\cite{RECO} summarized their performance in a fair way. In our experiment, we used CRAFT (Character-Region Awareness For Text detection)~\cite{CRAFT} for text detection and a text recognition model in~\cite{RECO}, because they show the state-of-the-art performance in those tasks.\par
In the experiment of Section~\ref{sec:exp}, word embedding is used to 
represent each word by a semantic vector. A pioneering work, called  Word2vec~\cite{WV}, is still a popular word embedding method. More recently, more elaborated word (or even sentence) embedding methods, such as BERT~\cite{BERT} and its versions~\cite{BERT-survey}, are proposed and utilized for various natural language processing tasks. Those methods will be useful for analyzing the semantics of scene texts. 
In~\cite{shinahara}, the words from scenery images were fed to word2vec and then their semantic vectors are classified into several clusters by k-means, in order to understand what kind of information is conveyed by scene texts. It is important to note that we can apply the word embedding methods not only to the words in scene texts but also to the object class names (as we see in Section~\ref{sec:visualize}).

%================================================================
\section{Co-Occurrence Analysis of Objects and Words}
%================================================================
%----------------------------------------------------------------
\subsection{Image Dataset}
%----------------------------------------------------------------
We used the training image set of OpenImages v4\footnote{\tt https://storage.googleapis.com/openimages/web/\\ index.html}  as a dataset to analyze the object-word co-occurrence.  Among 1,910,098 images in the dataset, 1,743,042 images with the bounding boxes of 600 predefined object classes are used; that is, the images with no object bounding box are discarded. The dataset contains 14,610,229 object bounding boxes in total; that is, on average, each image contains about eight object bounding boxes.
%----------------------------------------------------------------
\subsection{Scene Text Detection and Recognition}
%----------------------------------------------------------------
Scene texts in individual images were detected by using CRAFT~\cite{CRAFT}, which is one of the state-of-the-art scene text detectors based on CNN. CRAFT provides word bounding boxes. Each bounding box is rotatable and thus CRAFT can detect rotated texts. From 1,743,042 images, 7,364,937 word bounding boxes are detected. \par
The word bounding boxes were then fed to a state-of-the-art scene text recognizer proposed in~\cite{RECO}. In~\cite{RECO}, various combinations of functional modules are examined; among them, we used the combination ``TPS+ResNet+BiLSTM+Atten,'' where TPS (thin-plate spline) is employed for removing spatial distortion. ResNet, BiLSTM, and Atten (attention-based sequence prediction)  are used for feature extraction, sequence modeling, and the final word recognition.\par
\begin{figure}[t]
		\begin{center}
	\includegraphics[width=0.4\textwidth]{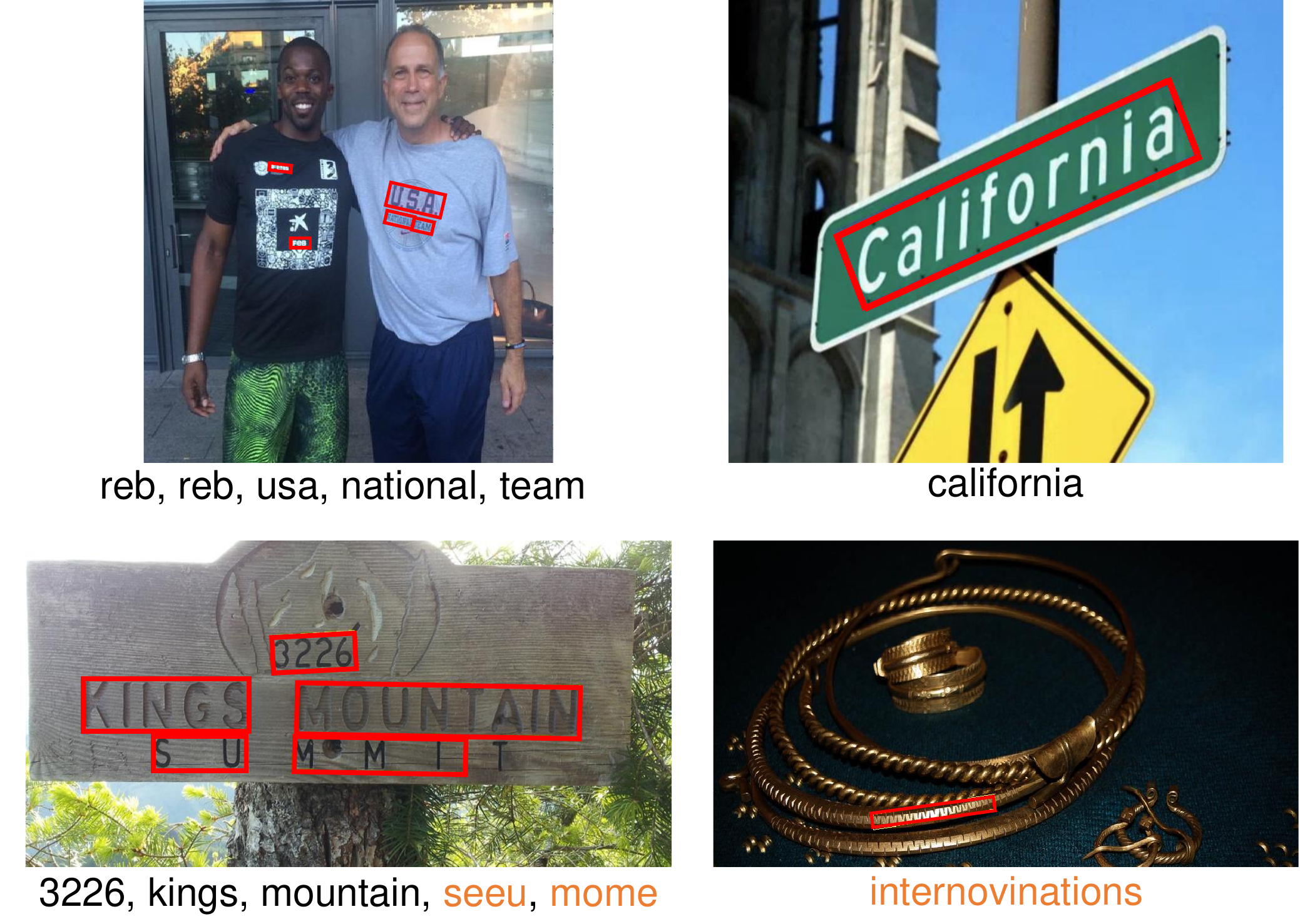}\\[-3mm]
    	\caption{Example of text detection and recognition results.}
    	\label{fig:EASTex}
    \end{center}
\end{figure}

Fig.~\ref{fig:EASTex} shows examples of scene text detection and recognition results. In the examples at the top row, 
even rotated texts are detected and recognized successfully. The examples at the bottom row show misrecognitions and 
a false detection\footnote{Since there is no ground-truth for text information in individual images, it is impossible to 
evaluate the detection and recognition accuracy. }. Note that most trivial misrecognitions (e.g., ``police'' as ``palice'') will be automatically corrected by the post-processing steps of Section~\ref{sec:postprocess}. Severe misrecognitions
will be automatically excluded from our analysis also by the post-processing. 

\begin{table*}[t]
\begin{center}
\caption{Top 40 frequent words in the OpenImages v4. The parenthesized number is the frequency rank in British National Corpus (BNC).}
%The underlined words are even not frequent (lower than $500$th rank) in BNC.}
\label{table:freq_words}\vspace{-3mm}
\begin{footnotesize}
\begin{tabular}{rl}
\hline
1-10th:&  new(26), one(3), \underline{army}(1325), may(37), police(530), man(47), go(1), good(12), free(314), stop(266), \\
11-20th:& world(82), use(20), state(121), day(29), open(136), time(10), city(188), would(4), get(6), big(108), \\
21-30th:& first(25), way(27), book(97), work(19), look(22), service(145), \underline{beer}(1968), home(87), school(75), \underline{bus}(722), \\
31-40th:& love(150), like(7), win(296), house(91), life(46), air(456), many(38), see(11), national(452), \underline{bar(747)} \\ \hline
\end{tabular}
\end{footnotesize}
\medskip
\caption{Top 40 frequent words in British National Corpus (BNC) dataset (for the comparison with Table~\ref{table:freq_words}).}
\label{table:freq_words-in-common}\vspace{-3mm}
\begin{footnotesize}
\begin{tabular}{rl}
\hline
% % before accepted
% 1-10th:& say, would, go, get, make, see, know, time, take, could,\\
% 11-20th:& year, think, come, last, give, people, also, well, new, may,\\
% 21-30th:& way, look, like, use, good, find, man, want, day, even,\\
% 31-40th:& many, one, yeah, thing, tell, back, still, must, child, put \\ \hline
% % after accepted
1-10th:& say, go, one, would, know, get, like, think, make, time, \\
11-20th:& see, good, people, year, take, well, come, could, work, use, \\
21-30th:& also, look, want, give, first, new, way, find, day, thing, \\
31-40th:& need, much, right, back, mean, even, may, many, last, child,  \\ \hline
\end{tabular}
\end{footnotesize}

\medskip
\caption{Top 40 objects with more words.}\label{table:object_freq_words}\vspace{-3mm}
\begin{footnotesize}
\begin{tabular}{rl}
\hline
1-10th:\!\!&\!\!\!\!\!\! whiteboard, poster, scoreboard, calculator, convenience store, parking meter, tablet computer, book, remote control, personal care, \\
11-20th:\!\!&\!\!\!\!\!\! tin can, mobile phone, ruler, bookcase, billboard, corded phone, ambulance, computer monitor, watch, ipod, \\
21-30th:\!\!&\!\!\!\!\!\! television, bus, refrigerator, box, telephone, perfume, wall clock, wine rack, camera, beer, \\
31-40th:\!\!&\!\!\!\!\!\! truck, plastic bag, office building, stop sign, computer keyboard, traffic sign, laptop, adhesive tape, bottle, van \\ \hline
\end{tabular}
\end{footnotesize}
\medskip
\caption{Top 40 objects with fewer words.}\label{table:object_less_freq_words}
\vspace{-3mm}
\begin{footnotesize}
\begin{tabular}{rl}
\hline
1-10th:& human ear, human eye, human mouth, footwear, bee, human nose, shower, hippopotamus, pomegranate, willow,  \\
11-20th:& duck, human face, goldfish, human beard, roller skates, kite, sparrow, goose, human hair, turkey,  \\
21-30th:& traffic light, swan, human leg, orange, flower, hedgehog, kangaroo, taco, lemon, marine invertebrates,  \\
31-40th:& human head, sea lion, shrimp, chair, porcupine, ladybug, red panda, pancake, glasses, bathroom accessory \\ \hline
\end{tabular}
\end{footnotesize}
\end{center}
\end{table*}

%----------------------------------------------------------------
\subsection{Counting Co-occurrence of Word and Object\label{sec:co-occur}}
%----------------------------------------------------------------
A word and an object are treated as co-occurring when 
the bounding box of the word is completely surrounded by the bounding box
of the object. In other words, if a word bounding box is partly overlapped with an object bounding box, they are not treated as co-occurring. 

%----------------------------------------------------------------
\subsection{Post-Processing Steps\label{sec:postprocess}}
%----------------------------------------------------------------
After text detection and recognition, several post-processing steps are applied to increase the reliability our co-occurrence analysis. The post-processing steps are organized in the following order. 
\begin{enumerate}
\item We removed the words whose bounding box is not surrounded by an object bounding box since we want to analyze the co-occurrence between texts and objects as noted in Section~\ref{sec:co-occur}.
\item The bounding boxes for stop-words are removed because stop-words will not show any strong correlation with specific objects. We used the stop-word list provided in Python NLTK Corpus.
\item The bounding boxes for the word ``I'' are also removed because we found that many edges in scenery images are wrongly detected as the word ``I.''
\item The recognition results are corrected by using the word2vec vocabulary\footnote{The word2vec vocabulary contains 1,111,684 words.} and the NGSL (New General Service List) vocabulary\footnote{The NGSL vocabulary contains 2,801 words including 91 stop-words.}.  
Specifically, if a recognized word is included in the word2vec vocabulary, we consider that the recognition result is correct. If not, 
the recognized word is replaced with the NGSL word which gives the minimum normalized Levenshtein distance. Note that, if the distance is larger than 0.15, the recognized word is discarded. Note that after this correction process the vocabulary size of the words in OpenImages becomes 30,182. 
\item Rare object-word co-occurrences are removed; specifically, if the number of the co-occurrences for an object-word pair is less 
than three, they are considered as unreliable co-occurrences and removed from the later analysis.
\end{enumerate}\par
Consequently, 19,543,919 word bounding boxes (equivalently, 19,543,919 object-word co-occurrences) were used in the analysis at Section~\ref{sec:exp}. They are comprised of 30,182 different words; this means that each word appears 64 times on average.\par
%----------------------------------------------------------------
\subsection{Word Embedding}\label{w2v}
%----------------------------------------------------------------
In several analyses in Section~\ref{sec:exp}, word embedding is used 
for vectorizing not only the detected words and the object class names (e.g., ``car''). Among various word embedding methods, word2vec~\cite{WV} is employed for representing each word as a 200-dimensional semantic vector. Although it is possible to use more recent methods, such as BERT~\cite{BERT}, word2vec is enough because our current target is a single word in general. Word2vec was trained by the standard 800 million-word corpus\footnote{{\tt https://github.com/hankcs/word2vec-google/blob/\allowbreak master/demo-train-big-model-v1.sh}}, collected from Wikipedia and UMBC corpus.\par

%================================================================
\section{Experimental Result\label{sec:exp}}
%================================================================
In this section, many analysis results will be shown using the 
co-occurrence of object-word pairs\footnote{\color{blue}The complete lists of the text-object co-occurences and the detected
text bounding boxes with their word recognition results are available
at https://github.com/Takeshiddd/Label-or-Message-A-Large-Scale-Experimental-Survey-of-Texts-and-Objects-Co-Occurrence.}. To clarify the analysis results, the following notations will be used:
\begin{itemize}
    \item $I$: the image set. $|I|=1,743,042$.
    \item $J$: the object class set. $|J|=600$.
    \item $K$: the word vocabulary. $|K|=30,183$.
    \item $h_{i,j,k}$: The number of times that the object $j\in J$ and the word $k\in K$ co-occur in the image $i\in I$.
    \item $f_{i,j}$: The number of the object $j$ in the image $i$.
    \item $g_{i,k}$: The number of the word $k$ in the image $i$.
\end{itemize}
In most cases, $h_{i,j,k}$, $f_{i,j}$, and $g_{i,k}$ are zero or one; by accumulating those values for all $i$ (e.g., $\sum_i h_{i,j,k}$), we can see various trends on the object-word co-occurrence. 

%----------------------------------------------------------------
\subsection{Frequent Words\label{sec:freq-word}}
%----------------------------------------------------------------
Table~\ref{table:freq_words} shows the 40 most frequent words, as one of the most primitive statistics of the words captured in the real images; specifically, this is 
the list of the words with the 40 largest $\sum_i g_{i,k}$. 
As suggested in ~\cite{shinahara}, the words captured in scenery images will have a different trend from the words in messages like Wikipedia and British National Corpus (BNC) dataset\footnote{{\tt http://www.kilgarriff.co.uk/bnc-readme.html}}. The parenthesized number in Table~\ref{table:freq_words}  is the frequency rank of the word in BNC (after the same post-processing steps, such as the stop-word removal). The underlined words, such as ``army,'' ``beer,'' and ``bus,'' are really frequent in the images but rather not frequent in BNC (below the 500th rank).\par
Table~\ref{table:freq_words-in-common} shows the 40 most frequent words in BNC (after stop-word removal). Since this corpus is comprised of various messages, this table shows the reliable word frequency in general messages. The clear difference between Tables~\ref{table:freq_words} and \ref{table:freq_words-in-common} indicates that the word frequency in the images is modified by the existence of non-message words, such as label words for some objects. In fact, our dataset contains various labels; in addition to the labels directly attached to objects, the alerts (e.g., ``stop'')  and the signs (e.g., ``open'') can be considered as labels for their surrounding environment.
%----------------------------------------
\begin{figure}[t]
\centering
\includegraphics[width=0.48\textwidth]{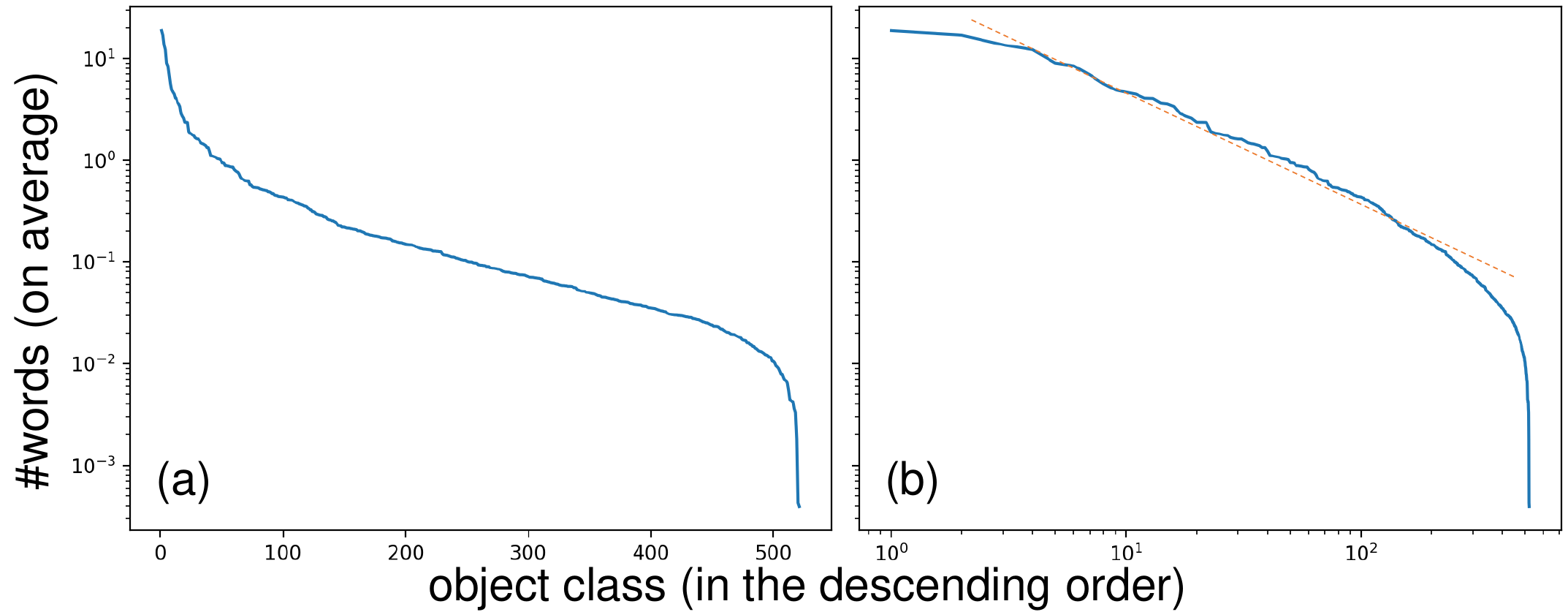}\\[-3mm]
\caption{The average number of words per object for each object class. (a)~Single and (b)~double logarithm representations. An orange straight line is superimposed to the graph of (b).}
\label{fig:average_words}
\end{figure}
%----------------------------------------
%----------------------------------------------------------------
\subsection{Objects with Many Co-Occurrences with Words}
%----------------------------------------------------------------
Table~\ref{table:object_freq_words} shows the objects with the 
40 most average co-occurrences with words. Specifically, the objects are ranked by the average number of words on each object, i.e., 
$(\sum_{i,k}h_{i,j,k})/(\sum_i f_{i,j})$. For the guarantee of analysis reliability, we excluded rare object classes whose appearing time in the dataset is less than 200, i.e.,  $\sum_i f_{i,j} < 200$; under this condition, 80 rare object classes are excluded.\par
Textual information is frequently attached to the surface or local parts (such as buttons) of the objects in Table~\ref{table:object_freq_words}. 
 For example, the top-ranked object ``whiteboard'' is a typical {\em message carrier} and many words telling some messages are written on it. An important observation is that almost all objects in Table~\ref{table:object_freq_words} are artificial. \par
Table~\ref{table:object_less_freq_words} shows the objects with fewer words. Clearly different from the objects in the previous table, most of the words in this table are natural. In fact, it is hard to find any textual information printed on a part of humans and animal bodies. Accordingly, the comparison between those tables leads to a simple conclusion: when we use textual information for more accurate recognition, artificial objects have more chance of improvement than natural objects.\par
Fig.~\ref{fig:average_words}~(a) shows the average number of words for each object, where the 520 (=600-80) object classes are arranged in the descending order of the average number. Only the top 50 objects (including the 40 objects of Table~\ref{table:object_freq_words}) put one or more words on their surface and the remaining objects put much less; the average word number of most objects lies between 0.5 and 0.01. This means the fine-grained object classification by using textual information often encounters the lack of words on the target objects. \par
Fig.~\ref{fig:average_words}~(b) reveals an interesting fact that the distribution of the average word number roughly follows Zipf's law (or, a power law). If a phenomenon follows this law, its distribution becomes linear in its double logarithm plot. It is well-known that the word frequency in a document (or a larger corpus) also follows Zipf's law (e.g.,~\cite{zipf}). Since the horizontal axis of our plot corresponds to not the ranked words but the ranked objects, we cannot directly apply the past linguistic theories about Zipf's law to ours. Future work might reveal the actual meaning of this fact.

\begin{table}[t]
\begin{center}
\caption{Top 40 object-word pairs with larger co-occurrence.}
\begin{tabular}{cccc|ccc}
\hline
rank & object   & word   &  & rank & object   & word       \\ \hline\hline
1    & poster   & new    &  & 21   & poster   & man        \\ \hline
2    & car      & police &  & 22   & book     & book       \\ \hline
3    & bus      & bus    &  & 23   & poster   & first      \\ \hline
4    & man      & army   &  & 24   & building & restaurant \\ \hline
5    & poster   & one    &  & 25   & poster   & get        \\ \hline
6    & clothing & army   &  & 26   & poster   & day        \\ \hline
7    & building & hotel  &  & 27   & building & shop       \\ \hline
8    & bus      & school &  & 28   & building & one        \\ \hline
9    & book     & new    &  & 29   & bottle   & beer       \\ \hline
10   & building & open   &  & 30   & poster   & good       \\ \hline
11   & building & bar    &  & 31   & bus      & city       \\ \hline
12   & book     & one    &  & 32   & building & house      \\ \hline
13   & building & new    &  & 33   & book     & may        \\ \hline
14   & person   & army   &  & 34   & airplane & air        \\ \hline
15   & poster   & world  &  & 35   & poster   & music      \\ \hline
16   & man      & new    &  & 36   & book     & would      \\ \hline
17   & poster   & may    &  & 37   & man      & one        \\ \hline
18   & poster   & time   &  & 38   & poster   & book       \\ \hline
19   & bus      & first  &  & 39   & beer     & beer       \\ \hline
20   & poster   & free   &  & 40   & poster   & like       \\ \hline
\end{tabular}
\label{table:Doujikakuritsu}
\end{center}
\end{table}

\begin{figure}[t]
\centering
\subfigure[obj=``car'', word=``police'']{
\includegraphics[width=0.23\textwidth]{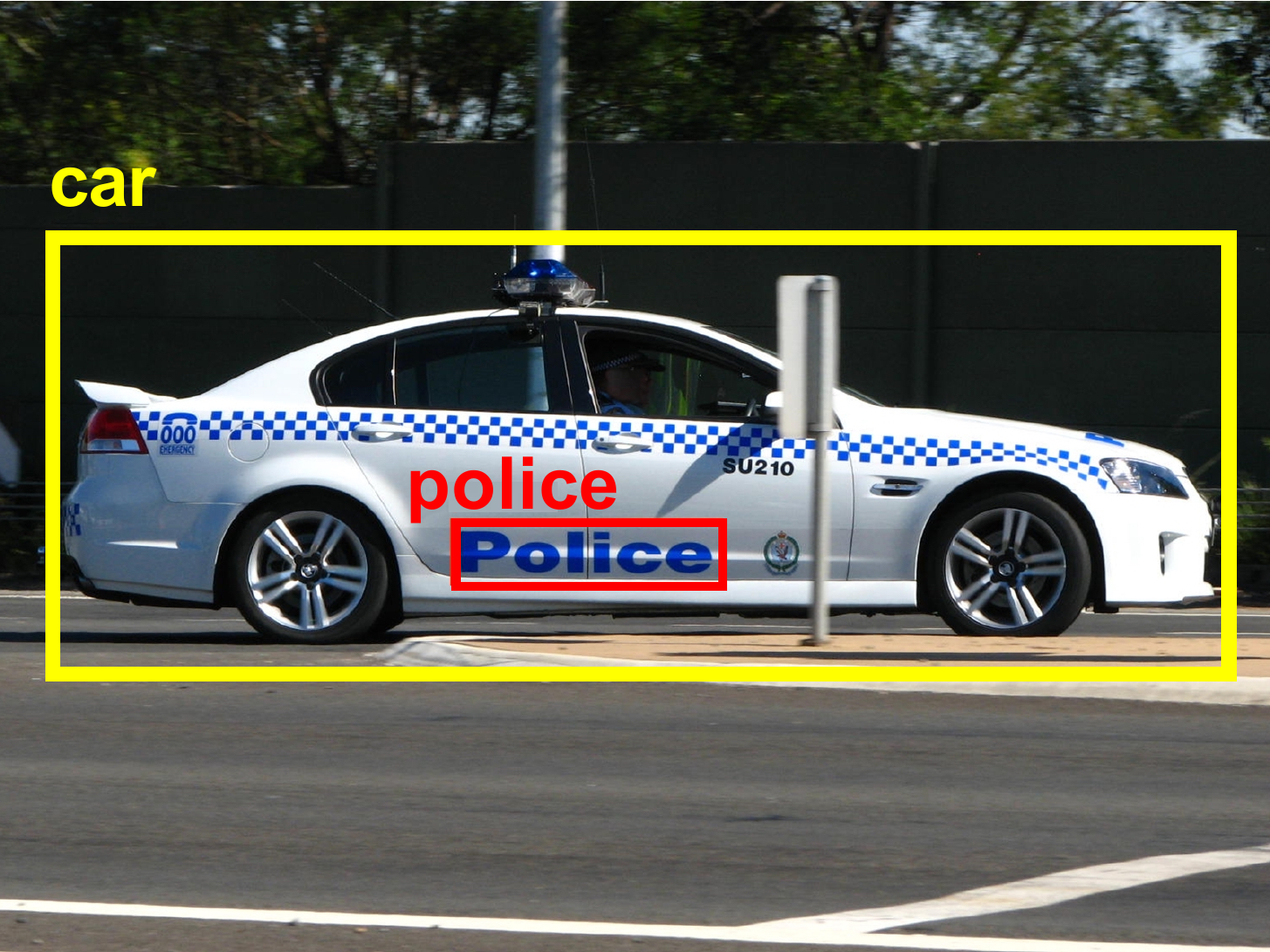}}
\subfigure[obj=``bus'', word=``bus'']{
\includegraphics[width=0.23\textwidth]{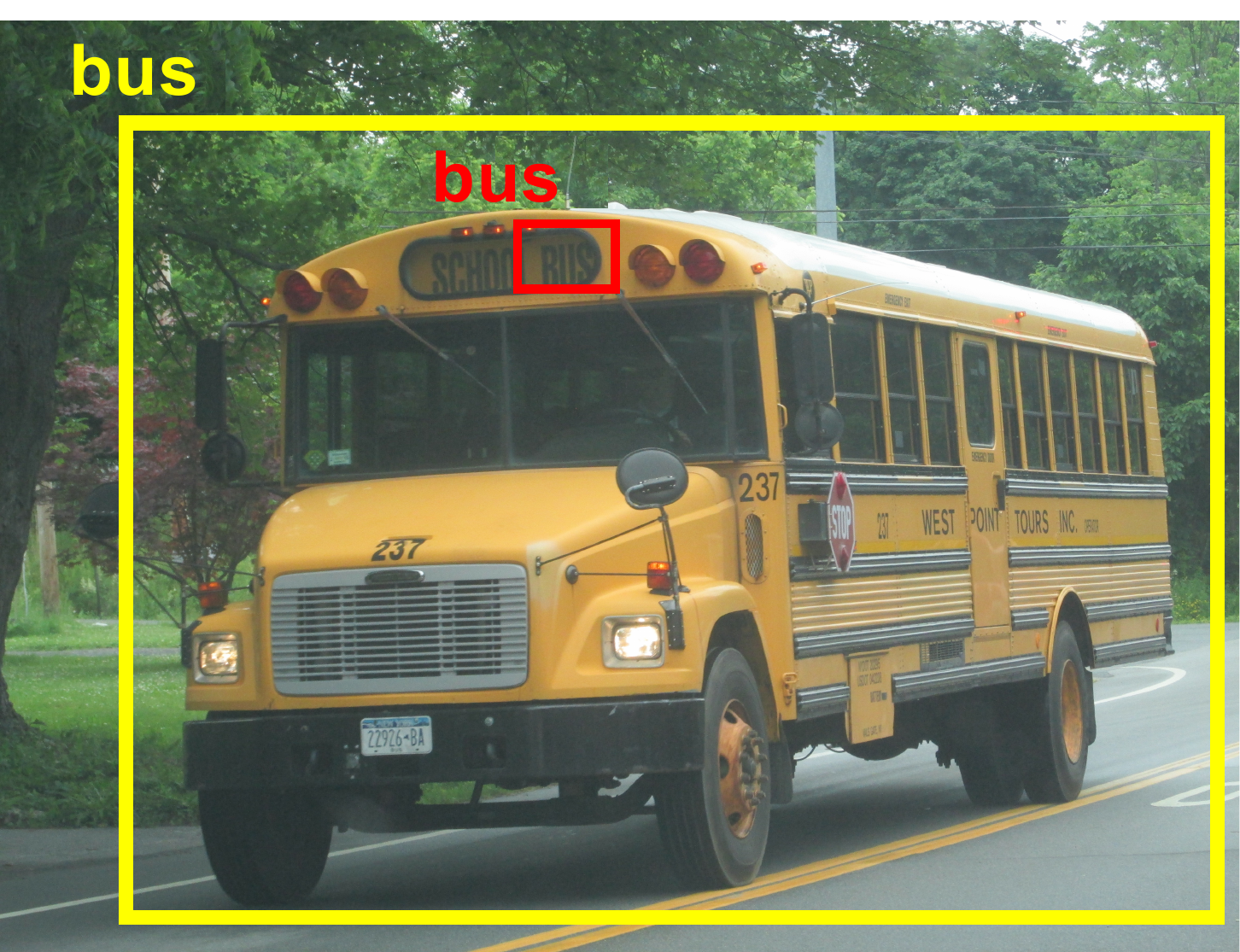}}\\[-3mm]
\caption{Examples of object-word pairs.}
\label{fig:word_object_pair}
\medskip
\includegraphics[width=0.5\textwidth]{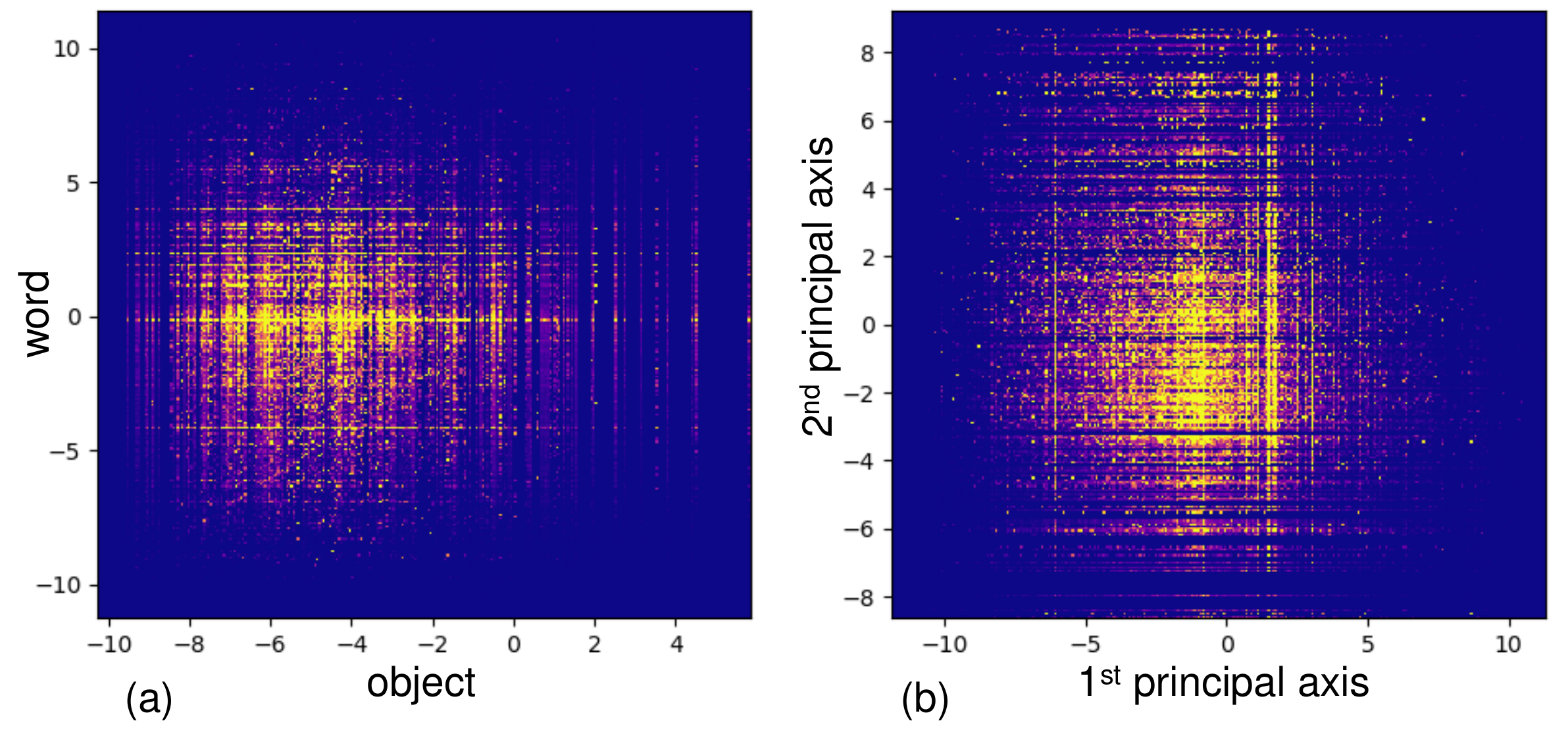}\\[-3mm]
\caption{Visualization of the object-word co-occurrence distribution by word2vec and PCA. (a)~Each of the 200-dimensional word spaces and the object space is projected into the 1-D principal axis independently and then 2-D product space is formed using the two obtained axes. (b)~The 400-dimensional object-word product space is projected into 2-D principal space.}
\label{fig:PCA}
\end{figure}

\begin{figure}[t]
		\begin{center}
	\includegraphics[width=0.48\textwidth]{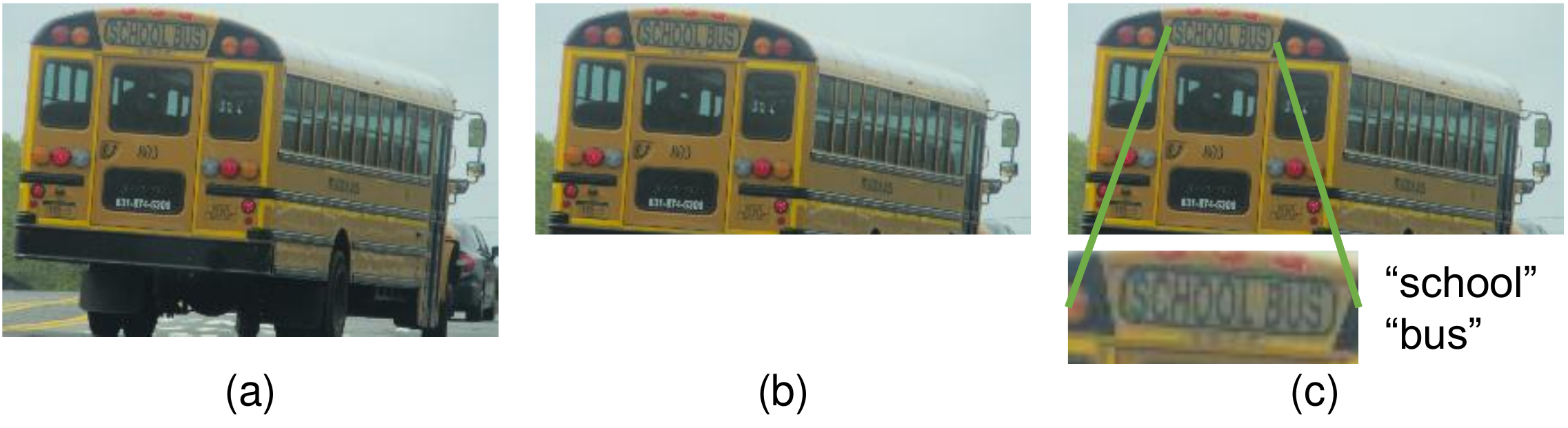}\\[-3mm]
    	\caption{Texts help object discrimination. (a)~A bus. (b)~Due to occlusion, the bus becomes like a train. (c)~Textual information ``bus'' and/or ``school'' helps to classify this object as ``bus.''}
    	\label{fig:bus}
    \medskip
	\includegraphics[width=0.48\textwidth]{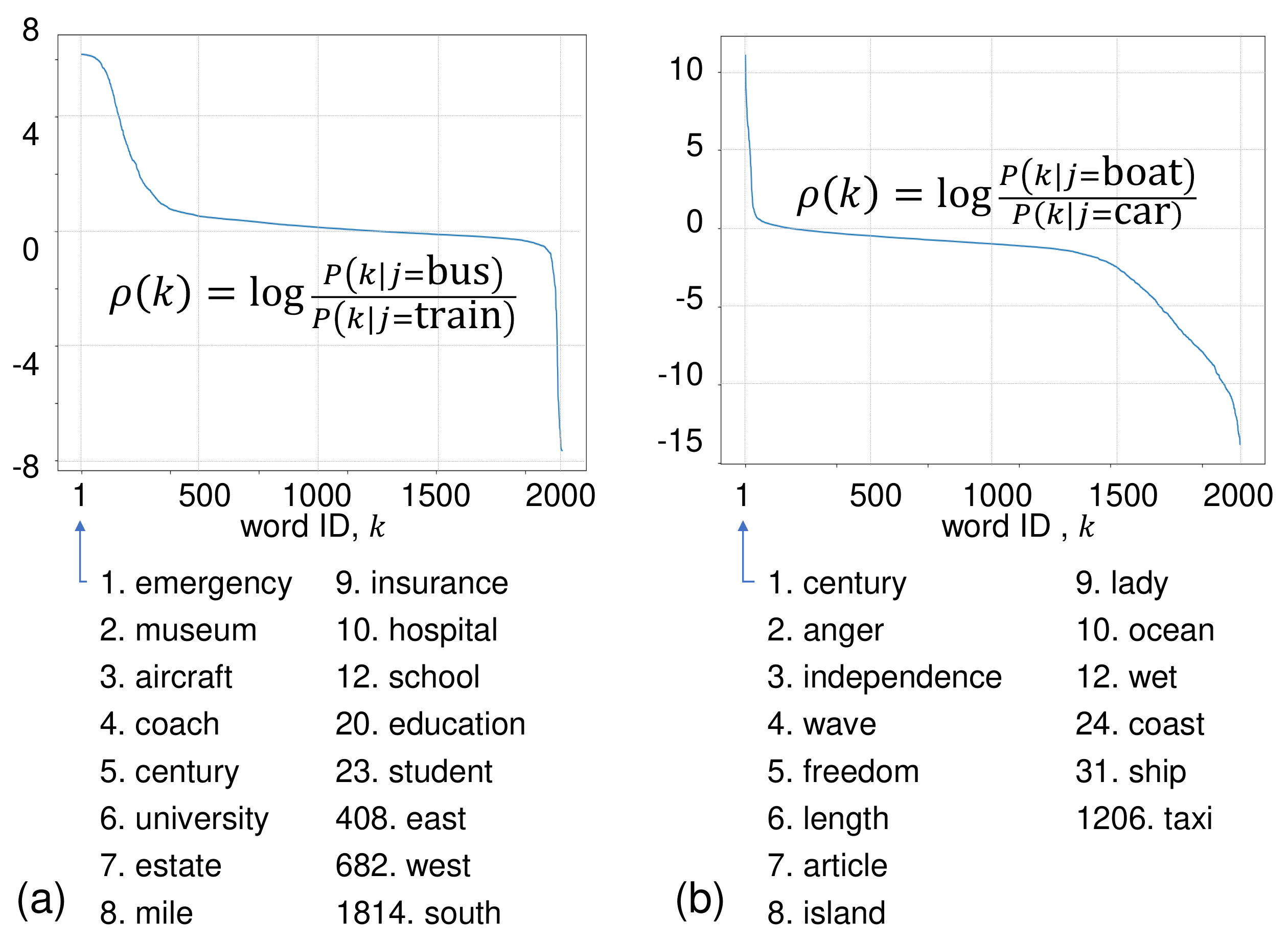}\\[-3mm]
    	\caption{Comparison of word frequency between two object classes by using the log probability ratio. (a)~``bus'' versus ``train.''  (b)~``boat'' versus ``car.''}
    	\label{fig:ratio}\vspace{-3mm}
    \end{center}
\end{figure}

%----------------------------------------------------------------
\subsection{Object-Word Pairs with Many Co-Occurrences}
%----------------------------------------------------------------
Table \ref{table:Doujikakuritsu} shows the object-word pairs with the 40 most co-occurrences;
that is, the $(j,k)$-pairs with the 40 largest $\sum_{i} h_{i,j,k}$.
For example, the object-word pair ``car''-``police'' has the second-largest co-occurrences; 
as shown in Fig.~\ref{fig:word_object_pair}~(a), the word ``police'' acts as the label for police cars. In Table \ref{table:Doujikakuritsu}, it is possible to find other words, such as ``hotel,'' ``bar,'' ``restaurant,'' and ``house'' for the object ``building.''  \par
The same object-word pairs, such as ``bus''-``bus'' at the third rank, are the typical case where the word is used as the label for disambiguating the object.  Fig.~\ref{fig:word_object_pair}~(b) shows this situation. As other examples, we can find ``book''-``book'' and ``beer''-``beer'' in this top 40 list. \par
Several object classes, such as ``poster'' and ``book'', appear frequently with abstract words, such as ``new'' and ``one''. In fact, these object classes will be message-carriers and thus sentences containing the abstract words are printed on them. This point will be further examined in Section~\ref{sec:message}.\par
%
%----------------------------------------------------------------
\subsection{Visualizing Object-Word Co-Occurrence by Word-Embedding and PCA\label{sec:visualize}}
%----------------------------------------------------------------
Fig.~\ref{fig:PCA} visualizes object-word co-occurrences not as a two-dimensional histogram but as a distribution in a semantic space. 
For this visualization, not only individual words but also object class names are converted into 200-dimensional semantic vectors by word2vec. 
We first have a 400-dimensional product space for representing word and object semantics at once. We then can 
visualize the semantic distribution of object-word pairs by applying standard PCA.  \par

Although we used two strategies on applying PCA, they result in similar distributions as shown in Figs.~\ref{fig:PCA}~(a) and (b). Specifically, in both figures, we can observe a grid-like distribution, rather than typical blob-like distributions. In other words, no intensive co-occurrence clusters exist in the distribution. The vertical lines forming the grid-like structure in (a) correspond to a set of objects where various words can be attached. This suggests that there are many message-carrying objects with various textual information; their existence will be proved experimentally by the entropy analysis in Section~\ref{sec:message}.
Similarly, the horizontal lines correspond to a set of words that can be attached to various objects. This suggests that there are many common words (like stop-words) over various objects; their existence is also proved in Section~\ref{sec:entropy}.

%----------------------------------------------------------------
\subsection{Difference of Word Co-Occurrences in Similar Objects}
%----------------------------------------------------------------
There is a possibility to discriminate confusing object classes by using the textual information attached to them. As an example, consider two confusing object classes, ``bus'' and ``train.'' The image appearance of these classes sometimes becomes very similar to each other. 
For example, the bus image shown in Fig.~\ref{fig:bus}~(a) becomes like (b) by partial occlusion, and then resembles a train.
If we find the word ``bus'', we can still recognize the object correctly as ``bus'', as shown in (c). This is simply because the word ``bus'' is not put to a ``train'' and thus the word acts as a very discriminative label. Similarly, if we find the word ``school'', the object is also recognized correctly as ``bus'' because there are no school-trains. In other words, if we find the difference of word co-occurrences in a pair of objects, we can expect that the words with different occurrences help to discriminate the objects.\par
Fig.~\ref{fig:ratio}~(a) confirms the fact that the textual information helps discrimination. This graph plots the log probability ratio:
$$
\rho(k)=\log \frac{P(k\mid j=\mathrm{bus})}{P(k\mid j=\mathrm{train})},
$$
by arranging the word $k$ in the descending order of $\rho(k)$. 
For better visibility of the plot, the vocabulary $K$ is limited to the frequent 2,000 NGSL words from the original 30,183 words, and then the smoothed co-occurrence $P(k\mid j)$ is calculated by a kernel density estimation:
$$
P(k\mid j)\propto \sum_{i,j',k'} h_{i, j’, k’}{\cal N}(v_k-v_{k'},w_j-w_{j'}), 
$$
where ${\cal N}$ is the two-dimensional Gaussian kernel. The vector $v_k$ is the semantic vector of the word $k$ given by word2vec and $w_j$ is the semantic vector of the class name of the object $j$.\par
Since the ratio is not always 0, there is a difference between the word occurrences of ``bus'' and ``train.'' Especially, the beginning part of the graph corresponds to the several words very specific to ``bus''; those words are ``emergency'', ``coach'', and ``university.'' The word $k$ around the middle part ($\rho(k)\sim 0$) will have no discriminative ability.
\par
Fig.~\ref{fig:ratio}~(b) shows another example between two objects, ``boat'' and ``car.'' Different from (a), the large ratio only appears the very beginning part of the graph. This means this pair has less discriminative words than (a). \par

\begin{table*}[t]
\begin{center}
\caption{Top 40 words with a higher entropy.}\vspace{-3mm}
\label{table:word_higher_entropy}
\begin{footnotesize}
\begin{tabular}{rl}
\hline
% % not front only
% 1-10th:&  red,  look, box,  go,  top,  bill,  cool, hi,  win, state,    \\
% 11-20th:&  use,  bell,  please,  ad,  low,  ear,  keep,  sky,  dog, pro, \\
% 21-30th:&   clear, touch, lot,  mail,  control,  green, ice,  loss, play, blue,  \\
% 31-40th:&   lake,  pen,  less,  thank,  ill,  unit,  real,  file, news, good \\ \hline
% % front only
1-10th:&  red,  go,  look, box,  use, state,  bill, lot,  low,  please,    \\
11-20th:&  top,  win,  start,  hi,   cool,  pro,   ice,   ad,  sweet,  lake, \\
21-30th:&  green,  bell, farm,  less,  blue,  control, ear,  touch, ill, hot,  \\
31-40th:&  keep,  mail,  water,  play,  dog,  light,  add,  mode,  sky,  clear\\ \hline
\end{tabular}
\end{footnotesize}
\medskip
\caption{Top 40 words with a lower entropy.}\vspace{-3mm}
\label{table:word_lower_entropy}
\begin{footnotesize}
\begin{tabular}{rl}
\hline
% % not front only
% 1-10th:& novel, author, hotel, fiction, restaurant, magazine, shift, pale, adult, policy, \\
% 11-20th:& army, entrance, taxi, pub, bus, telephone, story, complete, camera, modern, \\
% 21-30th:& exchange, another, popular, gallery, production, visit, volunteer, bottle, found, volume,\\ 
% 31-40th:& film, police, student, famous, performance, march, death, present, lunch,  money \\ \hline
% % front only
1-10th:& novel, author, fiction, restaurant, hotel, magazine, shift, pale, policy, entrance, \\
11-20th:& adult, pub, taxi, army, telephone , bus, story, exchange, camera, complete,  \\
21-30th:& modern, popular, another, gallery, volunteer, production, famous, bottle, police, visit, \\ 
31-40th:& film, march, performance, present, lunch, death, student, child, found, money \\ \hline
\end{tabular}
\end{footnotesize}
\medskip
\caption{Top 3 co-occurrence objects for the words with the 10 lowest entropy.}\vspace{-3mm}
\label{table:obj-for-low-entropy-word}
\begin{footnotesize}
\begin{tabular}{rcccccccccc}
\hline
% % not front only
% word:& novel& author& hotel& fiction& restaurant& magazine& shift& pale& adult& policy\\
% & [218] & [371] & [1111] & [328] & [754] & [250] & [265] & [385] &[210]& [389]\\  \hline
% top object: & poster & poster & building & poster & building & poster & computer keyboard & bottle &bus& man \\
% &(103)&(143)&(577)&(119)&(376)&(133)&(107)&(141)&(101)& (111)\\ 
% 2nd object: & book & book & house & book & house & book & laptop & beer &man& person \\
% &(79)&(130)&(171)&(112)&(112)&(23)&(56)&(113)&(19)& (101)\\ 
% 3rd object: & person & person & poster & bookcase & billboard & person & office supplies & drink &poster& suit \\
% &(9)&(11)&(75)&(20)&(51)&(14)&(30)&(25)&(12)& (38)\\ \hline
% front only
word:& novel& author& fiction& restaurant & hotel & magazine & shift& pale& policy& entrance\\
& [218] & [371] & [328]& [754]& [1111]   & [250] & [265] & [385] & [389] & [321] \\  \hline
top object: & poster & poster & poster & building & building & poster & computer keyboard & bottle & man &building \\
&(103)&(143)&(119)&(376)&(577)&(133)&(107)&(141)& (111)& (137)\\ 
2nd object: & book & book & book & house & house & book & laptop & beer & person & poster\\
&(79)&(130)&(112)&(112)&(171)&(23)&(56)&(113)&(101)& (44)\\ 
3rd object: & person & person  & bookcase & billboard & poster & person & office supplies & drink & suit & house\\
&(9)&(11)&(20)&(51)&(75)&(14)&(30)&(25)&(38)& (42)\\ \hline 

\end{tabular}
\end{footnotesize}
\medskip
\caption{Top 40 objects with a higher entropy.}\vspace{-3mm}
\label{table:object_higher_entropy}
\begin{footnotesize}
\begin{tabular}{rl}
\hline
1-10th:& poster, book, person, woman, man, clothing, plant, whiteboard, billboard, building,\\
11-20th:& girl, human body, land vehicle, computer monitor, mammal, table, shelf, bookcase, picture frame, house,\\
21-30th: & boy, office building, mobile phone, car, television, vehicle, box, window, convenience store, bottle,\\
31-40th: & auto part, human face, toy, laptop, suit, furniture, tablet computer, truck, drink, office supplies \\ \hline
\end{tabular}
\end{footnotesize}
\medskip
\caption{Top 40 objects with a lower entropy.}\vspace{-3mm}
\label{table:object_lower_entropy}
\begin{footnotesize}
\begin{tabular}{rl}
\hline
1-10th:& stop sign, cooking spray, ipod, ambulance, parking meter,  taxi,  coin, horse, juice,  remote control,  \\ 
11-20th:&  helicopter,  computer keyboard,  cat,  medical equipment,  calculator,  pen, corded phone,  cake,  drum,  helmet, \\
21-30th:& rose,  footwear,  dairy,  envelope, refrigerator,  doll,  tool,  stairs, waste container,  watercraft,\\
31-40th:& street light,  jeans, vehicle registration plate,  cart,  umbrella,  personal care, dessert,  fast food,  cosmetics,  airplane \\ \hline
\end{tabular}
\end{footnotesize}
\medskip
\caption{Top 3 co-occurrence words for the objects with the 10 lowest entropy.}\vspace{-3mm}
\label{table:word-for-low-entropy-obj}
\begin{footnotesize}
\begin{tabular}{rcccccccccc}
\hline
object:   & stop   & cooking & ipod  & ambulance & parking & taxi   & coin & horse & juice &  remote control  \\
          & sign   & spray   &       &           & meter   &        &      &       &       &  \\
          & [569]  & [839]  & [1401] & [1283]    & [1770]  & [1484] &[2838]& [1405]& [1767]&      [1160]  \\
top word: & stop   & oil     & menu  & service   & meter   & taxi   & trust&   net & orange&      menu  \\
          & (312)  & (59)   &(153)   & (50)      & (64)   & (118)   & (46) &  (55) &  (29) &      (30)  \\
2nd word: & cancer & stick   & music & emergency & hour    & duty   &  one&police  & protein& return  \\
          & (10)   &   (47)  & (21)  & (47)  & (52)  & (16)  &  (35)  &  (11)   &  (24)  &   (16)  \\
3rd word: & top    & less    & video & medical   & phone   & yellow &  god& park& sugar & enter  \\
          & (6)  & (25)  & (5)  & (17)  & (37)  & (14)  &(35)&(5)&(18)& (16)   \\ \hline 

\end{tabular}
\end{footnotesize}
\end{center}
\end{table*}

%----------------------------------------------------------------
\subsection{Entropy Analysis to Find Useful Words for Object Recognition\label{sec:entropy}}
%----------------------------------------------------------------
If a word appears only a limited number of object classes, we can use the word for more accurate object recognition; this is because we can utilize the existence of the word as a powerful prior for the classes. Conversely, if a word appears various objects, the word has no power to limit the object class and thus is not useful for object recognition.\par
To evaluate the variation of the objects for the word $k$, we use the conditional entropy: 
$$
    H_k(J) = -\sum_{j\in J} P_{j | k}\log P_{j | k},
$$
where $P_{j | k} = \sum_i h_{i,j,k}/\sum_{i,j} h_{i,j,k}$. The word $k$ whose object co-occurrence is less than 200 times (i.e.,  $\sum_{i,j}h_{i,j,k}<200$) is excluded from this analysis since its $P_{j | k}$ is unreliable.\par
Table~\ref{table:word_higher_entropy} shows the words with the 40 highest entropy $H_k(J)$. Many words in the list are adjectives and general verbs. They are, therefore, printed on various objects and not useful as a class prior for object recognition. (In other words, they are stop-words for object recognition.)\par
Table~\ref{table:word_lower_entropy} shows the word with the 40 lowest entropy $H_k(J)$. Each word in the table has a tight connection to specific object classes and therefore useful for object recognition. A detailed analysis is given in Table~\ref{table:obj-for-low-entropy-word}.  This table shows the top three co-occurring objects for the 10 lowest-entropy words.
The number in [$\cdot$] is $\sum_i g_{i,k}$, that is, the occurrence times of the word $k$. The number in ($\cdot$) is $\sum_i h_{i,j,k}$, that is, the co-occurrence times of the object-word pair. \par
Table~\ref{table:obj-for-low-entropy-word} indicates that there are three different cases that a word is useful for object recognition. The first case is the {\em strong-label word}.  An example in the table is ``hotel''; if this word is found on an object, the object can be recognized as ``building'' or ``house'' with a high probability. This is because ``hotel'' acts as a strong label of a building. The word ``restaurant'' also acts as a strong label. \par
The second case can be called a {\em weak-label word}. The words ``shift'' and ``pale'' are not a direct (i.e., strong) label of ``computer keyboard'' and ``bottle.'' The words, however, still suggest those objects in an indirect way. For example, ``shift'' is the label of a specific key and therefore, a weak label of ``computer keyboard.''\par

The third and promising case is {\em a word in a message}. Several words, such as ``novel,'' ``author,'' ``fiction,'' and ``magazine'', are mostly found on the objects ``poster'' and ``book.'' As we will see in Section~\ref{sec:message}, those objects are very typical message-carrying media. Therefore, the above words are shown on the objects {\em not as a label but as a part of the message}. This means that even a word in a message is sometimes effective for object recognition. The effectiveness comes from the difference of the vocabularies of the words captured in images and the words in general messages, as we showed in Tables~\ref{table:freq_words} and ~\ref{table:freq_words-in-common}.\par

%----------------------------------------------------------------
\subsection{Entropy Analysis to Find Message-Carrying Objects\label{sec:message}}
%----------------------------------------------------------------
We can assume that if the variation of the words on an object class
is very large, the object tends to be a message-carrier. In fact, various words will be written on the object ``whiteboard,'' which is a typical message-carrier. \par
To evaluate the variation of the words on the object $j$, we use the conditional entropy again, 
$$
H_j(K) = -\sum_{k\in K} P_{k | j}\log P_{k | j},
$$ 
where $P_{k | j} = \sum_i h_{i,j,k}/\sum_{i,k} h_{i,j,k}$. Similar to the previous section, the object $j$ whose word co-occurrence is less than 200 times (i.e.,  $\sum_{i,k}h_{i,j,k}<200$) is excluded from the analysis. If the entropy $H_j(K)$ becomes larger, the words on the object $j$ have more variations and the object $j$ will tend to be a message-carrier.\par
Table~\ref{table:object_higher_entropy} shows the objects with the 40 highest entropy $H_j(K)$. Many objects in the list can be treated as a message-carrier. ``Poster,'' ``book,'' ``computer monitor,'' and ``whiteboard'' are appropriate examples. Several object classes relating to persons are listed here because of word variations on their clothes --- therefore persons are also a message-carrier.\par
Conversely, if the variation is small, the object can limit the words on it. Table~\ref{table:object_lower_entropy} shows the objects with the 40 lowest entropy $H_j(K)$ and Table~\ref{table:word-for-low-entropy-obj} details the top 10 objects by showing their top-three co-occurring words. These objects are useful for recognizing words on them. In fact, if we find the object ``stop sign,'' the word on it will be ``stop'' with a high probability. Even though it is difficult for the object ``ambulance'' to determine the word on it, it is still possible to have a limited number of word candidates. Moreover, the candidates often have similar meanings to each other. For example, the words on the object ``ambulance'' are often relating to medical service.

%================================================================
\section{Conclusion}
%================================================================
The purpose of this paper is to conduct a large-scale analysis of object-word co-occurrence as an experimental survey of the relationship between visual objects and the textual information on the objects. The analysis results will be worthy of future research that tries to utilize the relationship for more accurate object recognition and text recognition and novel tasks such as scene-text VQA. In our analysis, we used 1.7 million real images from OpenImages v4 and a state-of-the-art scene text detector and a text recognizer. The resulting about 20 million object-word co-occurrences were then analyzed from various viewpoints.\par 
The main findings through our experimental surveys are summarized as follows:
\begin{itemize}
    \item The frequent words captured in real images show a different trend from the frequent words in the messages (from BNC corpus). This fact indicates that the word in the images often has an extra role of carrying a non-messages such as labels.
    \item Artificial objects have more chance to be helped by textual information for their accurate recognition results than natural objects because more words are attached to artificial objects.
    \item The average number of words per object is about 0.5$\sim$ 0.01 except for a limited number ($\sim 50$) of object classes and its distribution roughly follows Zipf's law.
    \item An entropy analysis reveals useful words for object recognition. Useful words act as a strong or weak label of the object. Even words in messages are sometimes useful because several words are very specific to message-carrying objects, such as ``poster'' and ``book.''
    \item Another entropy analysis reveals that the entropy can be a useful metric to discriminate the message-carrying objects from the others. It also shows the fact that the object class will limit the word vocabulary on the object.
\end{itemize}

Our future work will focus on the following tasks. First, we improve scene text detection and recognition by utilizing the knowledge derived from our analysis. One straightforward idea is to utilize the word information only if it is known that the word is useful as a positive prior to an object; if not, the word is not utilized for recognizing the object.
Second, we can relax the co-occurrence condition of Section~\ref{sec:co-occur}. Specifically, the relationship between  objects and spatially-distant texts is included as a new analysis target
because the label of an object is not always on it.
Third, we can apply the same analysis methodology to reveal the relationship between a place (i.e., environment) and textual information in the place. In fact, the word, such as ``danger'' on a signboard, can also be a label for the place (instead of a single object). For this purpose, we can utilize a very large scene image dataset (called places365) by Zhou~\cite{zhou} et al. 

\section*{Acknowledgment}
This work was partially supported by JSPS KAKENHI Grant Number JP17H06100.

% \begin{table*}[t]
% \begin{center}
% \caption{Top 40 object with a high probability of showing a word.}
% \begin{footnotesize}
% \begin{tabular}{rl}
% \hline
% 1-10th:& book, keyboard, bus, man, sign, beer, clock, poster, airplane, can,\\
% 11-20th:& camera, car, clothing, plate, person, computer, girl, vehicle, mobile, drink,\\
% 21-30th:& building, wine, phone, tree, snack, tin, registration, calculator, laptop, store,\\
% 31-40th:& body, house, coin, land, human, juice, billboard, meter, parking, scoreboard\\ \hline 
% \end{tabular}
% \label{table:prob_sort}
% \end{footnotesize}
% \end{center}
% \end{table*}

\end{document}